\documentclass[11pt]{article}

\usepackage[margin=1in]{geometry}
\usepackage{graphicx}
\usepackage{amsmath, amssymb}
\usepackage{booktabs}
\usepackage{subcaption}
\usepackage{float}
\usepackage{array}
\usepackage[font=small,labelfont=bf]{caption}
\usepackage[colorlinks=true,linkcolor=blue,citecolor=blue,urlcolor=blue]{hyperref}

\newcolumntype{C}[1]{>{\centering\arraybackslash}p{#1}}

\title{Sim-to-Real Fruit Detection Using Synthetic Data: Quantitative Evaluation and Embedded Deployment with Isaac Sim}

\author{
Martina Hutter-Mironovova\\
YASKAWA Europe GmbH\\
\texttt{martina.hutter@yaskawa.eu}
}

\date{}

\begin{document}

\maketitle

\begin{abstract}
This study investigates the effectiveness of synthetic data for sim-to-real transfer in object detection under constrained data conditions 
and embedded deployment requirements. Synthetic datasets were generated in NVIDIA Isaac Sim and combined with limited real-world fruit images 
to train YOLO-based detection models under real-only, synthetic-only, and hybrid regimes. Performance was evaluated on two test datasets: 
an in-domain dataset with conditions matching the training data and a domain shift dataset containing real fruit and different background conditions. 
Results show that models trained exclusively on real data achieve the highest accuracy, while synthetic-only models exhibit reduced performance 
due to a domain gap. Hybrid training strategies significantly improve performance compared to synthetic-only approaches and achieve results close 
to real-only training while reducing the need for manual annotation. Under domain shift conditions, all models show performance degradation, 
with hybrid models providing improved robustness. The trained models were successfully deployed on a Jetson Orin NX using TensorRT optimization, 
achieving real-time inference performance. The findings highlight that synthetic data is most effective when used in combination with real data 
and that deployment constraints must be considered alongside detection accuracy.
\end{abstract}


\section{Introduction}

Recent development in artificial intelligence has significantly expanded capabilities of automation, enabling the reliable execution of tasks 
that were previously difficult or costly to automate. In particular, the rapid development of deep neural networks and accessible machine learning 
frameworks has facilitated the deployment of perception systems in dynamic industrial environments. These models can be adapted to specific applications 
or fine-tuned from pre-trained networks, allowing efficient development even with limited task-specific data. Object picking, sorting and quality 
inspection represent such applications. Due to variability in shape, color, illumination, and occlusions, as well as low batch, high mix manufacturing, 
these tasks are challenging for traditional automation and machine vision approaches, whereas AI-based perception methods offer a robust and 
scalable alternative for a challenging automated processing.

Reliable and fast object detection is a fundamental component of automated fruit handling systems, where accurate perception directly influences grasp planning algorithm, 
cycle time, and overall system robustness. In industrial and semi-structured environments, fruit detection must operate under varying lighting 
conditions, background clutter and partial occlusions. Deep learning–based object detectors have significantly improved visual perception capabilities; 
however, their performance depends heavily on the availability of annotated training data.

Collecting and labeling large-scale real-world datasets for fruit detection can be time-consuming and costly. Variations in fruit shape and appearance, 
pose, lighting conditions, and scene configuration require extensive sampling to achieve robust generalization. In practical automation settings, 
only a limited number of labeled real images may be available during system integration. This constraint motivates the use of synthetic data generation 
as a scalable alternative for training perception models.

Modern simulation platforms such as NVIDIA Isaac Sim enable photorealistic rendering and automatic annotation of synthetic scenes. Through domain 
randomization, scene parameters including lighting, camera pose, background textures, and object placement can be systematically varied to enlarge 
the dataset and improve model robustness. Synthetic data offers several advantages, i.e. fully automated data labeling, controlled variability, 
and reproducibility. However, a challenge that remains is the sim-to-real gap - the performance degradation that occurs when models 
trained on simulated data are applied to infer on real-world images.

Previous research has shown that synthetic datasets and domain randomization can enable effective sim-to-real transfer for perception 
models \cite{bonetto-2023}, \cite{shing-2024}. However, several practical questions remain insufficiently quantified in industrial contexts. 
In particular, it is unclear how much synthetic data is required to meaningfully reduce the need for real training images, and to what extent such 
models retain performance when deployed on embedded edge hardware commonly used in automation systems. While embedded deployment of deep learning 
models has been widely studied \cite{soliman-2026}, \cite{kareemah-2025}, the combined analysis of synthetic training strategies and real-time 
inference constraints remains limited.

Embedded GPU platforms such as the Jetson Orin NX provide substantial computational capability within strict power and latency constraints, 
making them suitable for real-time perception in automation cells. However, deployment requires model optimization, typically through frameworks 
such as TensorRT, and performance must be evaluated under realistic operating conditions. Accuracy improvements achieved during offline training 
are only meaningful if they can be preserved under embedded deployment constraints.

This paper presents a quantitative analysis of sim-to-real transfer for fruit detection using synthetic data generated in Isaac Sim. A lightweight 
object detection model, YOLOv8s, is trained under real-only, synthetic-only, and hybrid training regimes. Two synthetic dataset sizes 
(1000 and 5000 synthetic images) are evaluated to analyze the impact of synthetic scale. Experiments are conducted using limited real training datasets 
(50 and 100 real images) and two fixed real-world test datasets (in-domain and domain shift), each consisting of 100 images. In addition to detection accuracy metrics, the trained models are exported 
to ONNX and optimized with TensorRT for deployment on the Jetson Orin NX 16GB platform, where end-to-end inference latency and throughput are measured.

The main contributions of this work are as follows:
\begin{itemize}
\item	A systematic comparison of real-only, synthetic-only, and hybrid training strategies for fruit detection under limited real-data conditions.
\item	A quantitative evaluation of the effect of synthetic dataset size on sim-to-real transfer performance.
\item	An embedded deployment study demonstrating real-time inference on a Jetson Orin NX 16GB using a TensorRT-optimized pipeline.
\item	Practical insights regarding the amount of synthetic and real data required for edge-ready perception in automation environments.
\end{itemize}

The presented results provide insight into how synthetic data can be leveraged to reduce data collection effort while maintaining real-world 
detection performance and meeting embedded deployment constraints.

\section{State of the Art}
\subsection{Synthetic Data for Object Detection}

The use of synthetic data for training deep learning models has gained significant attention due to the  reduction of manual data collection and 
annotation effort and improved scalability \cite{nikolenko-2021}. Simulation environments enable automatic generation of labeled datasets with full 
control over the whole scene composition, lighting conditions, placement of objects, and camera parameters \cite{tremblay-2018}. Modern platforms such as 
NVIDIA Isaac Sim \cite{isaacsim}, built on NVIDIA Omniverse, provide photorealistic object rendering and automated annotation pipelines for object detection, 
segmentation, and pose estimation tasks.

Two principal strategies are commonly used in synthetic data generation. The first emphasizes photorealism to closely approximate real-world 
appearance of objects. The second strategy, known as domain randomization, introduces variability in lighting, surface textures, 
object poses, and backgrounds variations to encourage invariance to visual differences between simulated and real domains. Domain randomization 
improves generalization by exposing the model to a broad distribution of visual conditions during training \cite{tobin-2017}. However, purely synthetic 
datasets often suffer from residual domain discrepancies, which leads to degraded real-world 
performance \cite{sadeghi-2017}, \cite{peng-2018}, \cite{tremblay-2018}.

Despite extensive research on synthetic data generation, there are fewer studies that provide systematic evaluations of to what extent synthetic dataset 
size influences real-world detection performance under constrained real-data scenarios \cite{nikolenko-2021}, \cite{hattori-2015}. 
In industrial contexts, where only limited amount of real datasets are available during system integration, understanding the quantitative 
trade-off between synthetic scale and real data requirements remains essential.

\subsection{Sim-to-Real Transfer in Robotics}

The sim-to-real problem refers to the differences in performance observed when models, that are trained in simulation environment, are deployed in 
real-world environment \cite{tobin-2017}, \cite{peng-2018}. Differences in sensor noise, lighting conditions, material properties, and background 
complexity contribute to this gap. In robotics research, sim-to-real transfer has been studied extensively in manipulation, navigation, and 
reinforcement learning tasks \cite{james-2019}, \cite{muratore-2022}, while early work focused primarily on control policies, visual perception has 
become an increasingly important area of investigation.

Hybrid training strategies, in which synthetic pretraining is followed by fine-tuning on a limited set of real images, are frequently reported 
to outperform both synthetic-only and real-only approaches \cite{tremblay-2018}. Synthetic data can provide coverage of diverse geometric 
configurations and environmental variations, while real images help anchor the model to actual sensor characteristics \cite{hattori-2015}. 
Nevertheless, the amount of real data required to stabilize transfer and the marginal benefit of increasing synthetic dataset size remain 
open empirical questions in many application domains \cite{nikolenko-2021}.

In perception-driven automation tasks such as fruit handling, sim-to-real studies are comparatively limited. Many works evaluate detection 
accuracy in agricultural field settings; however, fewer focus on controlled industrial environments and embedded deployment 
scenarios \cite{koirala-2019}, \cite{sa-2016}. A systematic comparison of real-only, synthetic-only, and hybrid regimes under fixed real test 
conditions is therefore valuable for quantifying practical sim-to-real effectiveness.

\subsection{Deep Learning–Based Fruit Detection}

Deep convolutional neural networks have been used for fruit detection in both agricultural and industrial 
applications \cite{koirala-2019}, \cite{sa-2016}. Single-stage object detectors from the YOLO family have gained their popularity due to their 
trade-off between accuracy and inference speed. Their variations, such as YOLOv5 \cite{glenn-2020} and YOLOv8 \cite{jocher-2023} provide 
efficient architectures, that are suitable for real-time object detection.

In fruit detection tasks, YOLO-based models have been applied to variety of tasks from orchard monitoring, harvest estimation, fruit maturity
and robotic harvesting systems \cite{chen-2021}, \cite{yu-2020}. Such studies typically emphasize detection accuracy under varying environmental 
conditions. However, large datasets are often required to achieve robust performance of the system, and training from scratch is rarely feasible 
in industrial deployments with limited data availability \cite{sun-2017}.

Lightweight model variants (e.g., nano or small configurations) are particularly relevant for embedded applications, where computational and 
power constraints limit model complexity \cite{zhao-2022}. Although several works report accuracy metrics, fewer analyze performance in low-data 
regimes combined with synthetic augmentation strategies. Additionally, embedded deployment metrics such as end-to-end latency and power-aware 
inference performance are often omitted \cite{redmon-2018}.

\subsection{Embedded Edge Deployment of Object Detectors}

Real-time perception in automation systems frequently relies on embedded GPU platforms \cite{lane-2015}, \cite{shi-2016}. The Jetson Orin NX 
represents a high-performance embedded solution capable of executing deep neural networks within constrained power envelopes \cite{nvidiajetson}. 
Deployment typically involves model export to ONNX format followed by optimization using TensorRT, which enables precision 
reduction (e.g., FP16 or INT8) and kernel-level optimizations \cite{tensorrt}.

While numerous studies benchmark object detectors on powerful desktop GPUs, comparatively fewer report detectors performance on embedded hardware 
under documented power configurations. In industrial automation, reporting only offline accuracy is insufficient. Practical deployment requires 
evaluation of latency, throughput, and computational utilization under realistic constraints. End-to-end latency—including preprocessing, inference, 
and postprocessing—directly impacts system cycle time and project feasibility \cite{zhao-2022}.

Existing literature often treats accuracy improvement and embedded deployment as separate topics. A comprehensive evaluation that links sim-to-real 
transfer effectiveness with edge inference performance remains limited. Integrating both aspects provides a more complete assessment of whether 
synthetic data strategies are viable for real-world automation systems \cite{reddi-2020}.

\subsection{Research Gap}

In summary, prior work has demonstrated the potential of synthetic data, domain randomization, and lightweight object detectors for robotic perception. 
However, limited research quantitatively analyzes: 
\begin{enumerate}
\item    the impact of synthetic dataset size under small real-data regimes, 
\item    the comparative performance of real-only, synthetic-only, and hybrid training strategies on a fixed real test set, and 
\item    the preservation of detection accuracy after deployment on embedded GPU hardware.
\end{enumerate}

The present study addresses this gap by systematically evaluating sim-to-real transfer for fruit detection using synthetic data generated 
in Isaac Sim and by benchmarking the resulting models on an embedded edge platform under documented deployment conditions.

\section{Methodology}
\subsection{Overview}

This study evaluates the effect of synthetic data on sim-to-real transfer for fruit detection under constrained real-data conditions and embedded 
deployment requirements. The methodology consists of three main components: 
\begin{enumerate}
\item    synthetic dataset generation in simulation, 
\item    preparation of real-world datasets, and 
\item    systematic training under real-only, synthetic-only, and hybrid regimes. 
\end{enumerate}

All experiments were conducted using a fixed evaluation protocol to ensure comparability.

The target object classes considered in this study are:
\begin{itemize}
\item	apple\_red
\item	apple\_green
\item	banana
\item   grapes
\item   lemon
\item   orange
\item   strawberry
\end{itemize}

In addition to the target object classes (apple\_red, apple\_green, banana, grapes, lemon, orange, and strawberry), both synthetic and real-world 
datasets included non-target objects acting as visual distractors. These distractors were not annotated and were intentionally introduced to 
increase scene complexity and better reflect real-world environments, where irrelevant objects are commonly present. This setup enables evaluation 
of the model's robustness to clutter and reduces the risk of overfitting to simplified scenes containing only target classes.

\subsection{Synthetic Dataset Generation in Isaac Sim}

Synthetic data were generated using NVIDIA Isaac Sim. The simulation environment was configured to approximate an industrial fruit handling scenario, 
including a planar working surface and camera-based perception setup.

\begin{figure}[H]
\centering
\begin{subfigure}{0.48\textwidth}
\includegraphics[width=\linewidth]{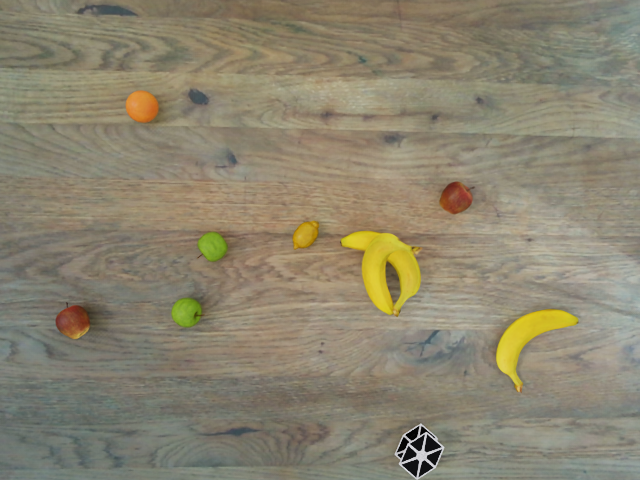}
\end{subfigure}
\begin{subfigure}{0.48\textwidth}
\includegraphics[width=\linewidth]{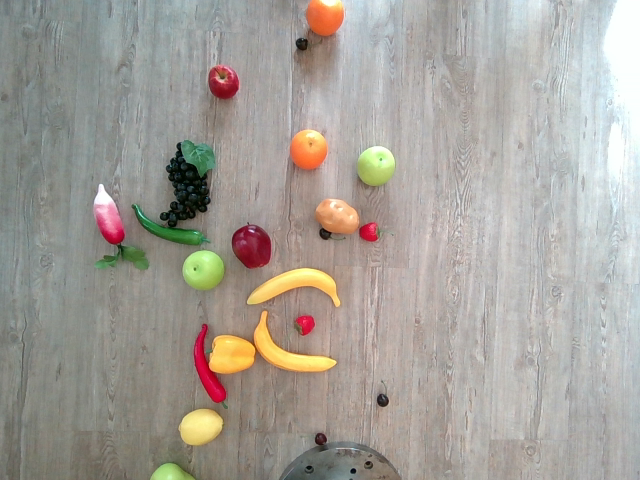}
\end{subfigure}
\caption{Comparison between synthetic and real-world data. 
(a) Example image generated in NVIDIA Isaac Sim. 
(b) Real-world image from the train dataset. 
Differences in texture, lighting, and object appearance illustrate the domain gap between simulated and real environments.\label{fig1}}
\end{figure}

Figure~\ref{fig1} illustrates the visual differences between synthetic and real-world data. While the simulated environment provides controlled 
object placement and consistent geometry, discrepancies in texture realism, lighting conditions, and surface properties introduce a domain gap 
that impacts model generalization.

\subsubsection{Scene Configuration}

The simulated scene consisted of:
\begin{itemize}
\item	One or more fruit objects per frame
\item	A supporting surface (i.e., table)
\item	A static RGB camera mounted above the workspace
\end{itemize}

Camera intrinsics (focal length, principal point) were fixed during generation, while extrinsics were slightly randomized to simulate realistic 
viewpoint variation.

\subsubsection{Domain Randomization}

To improve generalization to real-world conditions, domain randomization was applied to the following parameters:
\begin{itemize}
\item	Lighting: intensity, direction, and color temperature
\item	Object pose: random translation and rotation within the workspace
\item	Object count: single and multiple fruit instances
\item	Partial occlusions: controlled overlap between objects
\item	Camera pose jitter: small perturbations in position and orientation
\end{itemize}

Randomization ranges were defined prior to dataset generation and kept constant across experiments. A fixed random seed was used to ensure 
reproducibility.

\subsubsection{Dataset Size and Annotation}

Two synthetic dataset sizes were generated:
\begin{itemize}
\item	S1000: 1000 synthetic images
\item	S5000: 5000 synthetic images
\end{itemize}

All images were rendered at a fixed resolution and automatically annotated using simulator-generated 2D bounding boxes. The annotation process 
required no manual intervention. Synthetic datasets were divided into training and validation splits using a fixed ratio.

Only target classes were annotated; distractor objects were ignored during training and evaluation.

\subsection{Real-World Dataset}

A real-world fruit dataset was collected using an RGB camera under laboratory conditions representative of a small automation cell. The dataset was 
manually annotated with 2D bounding boxes.

The dataset was divided as follows:
\begin{enumerate}
\item    Training subsets:
	\begin{itemize}
	\item	R50: 50 images
	\item	R100: 100 images
	\end{itemize}

\item   Test datasets: 
	\begin{itemize}
	\item	T1-100: 100 images with conditions consistent with the training data (plastic fruit, same background), fixed and not used for training or hyperparameter tuning;
	\item 	T2-100: 100 images representing a domain shift scenario (real fruit and different background), fixed and not used for training or hyperparameter tuning.
	\end{itemize}
\end{enumerate}

The test set remained constant across all experiments to ensure fair comparison between training regimes.

\subsection{Object Detection Model}

All experiments were performed with YOLOv8 object detection architecture, implemented through the Ultralytics framework. Specifically, the 
YOLOv8s (small) variant was used and initialized from weights pretrained on a large-scale object detection dataset. The YOLOv8 family belongs to 
the class of single-stage detectors and is designed to provide a favorable balance between detection accuracy and inference speed.

The small configuration was selected as a compromise between model capacity and computational efficiency. Compared to the nano variant, YOLOv8s 
offers improved detection accuracy while still maintaining inference performance suitable for embedded GPU platforms. This makes it an appropriate 
choice for perception tasks deployed in edge computing environments.

The same model architecture, pretrained initialization, and training configuration were used across all experiments to ensure a consistent 
comparison between the evaluated training regimes.

\begin{figure}[H]
\centering
\includegraphics[width=10.0 cm]{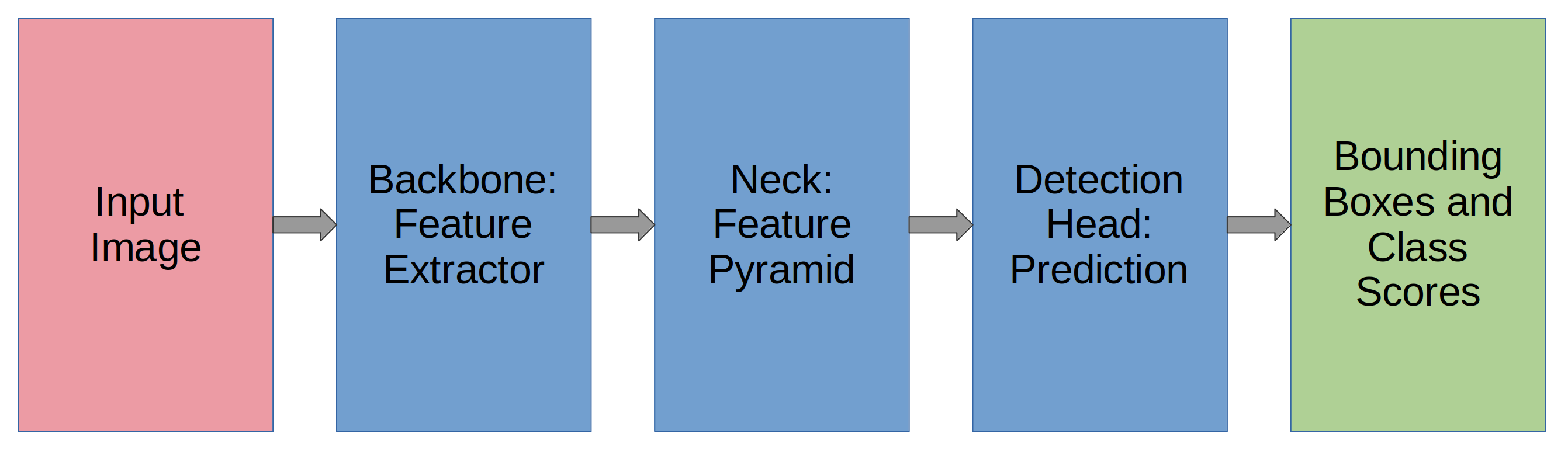}
\caption{General architecture of a YOLO-based object detection network. The model consists of a backbone for feature extraction, a neck for 
multi-scale feature aggregation, and a detection head that predicts bounding boxes and class probabilities in a single forward pass. 
Adapted from Redmon and Farhadi \cite{redmon-2018}.\label{fig2}}
\end{figure}

The YOLOv8 architecture follows a typical single-stage detector design consisting of three main components: a backbone, a neck, and a detection 
head (Figure~\ref{fig2}). The backbone extracts hierarchical image features through a series of convolutional layers, while the neck aggregates 
multi-scale feature maps using a feature pyramid structure to improve detection of objects at different spatial scales. The detection head 
predicts bounding box coordinates, objectness scores, and class probabilities in a single forward pass simultaneously. This unified architecture 
enables efficient inference and makes YOLO-based detectors particularly suitable for real-time perception systems and embedded edge deployment.

\unskip

\subsection{Training Regimes}

To quantify the contribution of synthetic data, multiple training regimes were defined. All models were trained using identical hyperparameters, 
including learning rate schedule, optimizer configuration, batch size, input resolution, and number of epochs. No regime-specific tuning was performed.

The following configurations were evaluated:
\subsubsection{Real-Only Training}
\begin{itemize}
\item	R50: 50 real images
\item	R100: 100 real images
\end{itemize}
These serve as baselines representing low-data industrial scenarios.

\subsubsection{Synthetic-Only Training}
\begin{itemize}
\item	S1000: 1000 synthetic images
\item	S5000: 5000 synthetic images
\end{itemize}
These experiments evaluate the direct sim-to-real transfer without real fine-tuning.

\subsubsection{Hybrid Training}
\begin{itemize}
\item	H1000+50: 1000 synthetic + 50 real images
\item	H5000+50: 5000 synthetic + 50 real images
\item   H1000+100: 1000 synthetic + 100 real images
\item   H5000+100: 5000 synthetic + 100 real images
\end{itemize}
In hybrid configurations, synthetic and real samples were combined within the same training process. The ratio between synthetic and real samples 
was preserved across epochs.

\subsubsection{Training Configuration}

All models were trained using the Ultralytics YOLOv8s implementation on a GPU-enabled system with CUDA support.

Baseline models (real-only and synthetic-only) were trained using an input resolution of 832×832 pixels, a batch size of 24, and 150 epochs. Early stopping was disabled by setting the patience parameter to zero. Default Ultralytics training hyperparameters were used.

Training datasets were split into training and validation subsets using a 70/20 ratio. The validation set was used for model selection, and the 
best-performing checkpoint was determined based on validation mAP@0.5. Separate fixed test datasets (T1-100 and T2-100) were used 
exclusively for evaluation and were not involved in training or model selection.

Hybrid models were trained using a fine-tuning strategy. Models pretrained on synthetic datasets were further trained on real-world data for an 
additional 50 epochs using the same configuration, enabling adaptation to real-world conditions.

All models were initialized from COCO-pretrained weights provided by the Ultralytics framework.

During evaluation, consistent default inference parameters from the Ultralytics framework were applied across all experiments.

Experiments were conducted using Ultralytics version 8.4.14 and PyTorch 2.10 on a CUDA-enabled GPU.

Default YOLO augmentation settings were used.

\subsection{Evaluation Protocol}

All trained models were evaluated on two fixed real-world test datasets (T1-100 and T2-100), each consisting of 100 images. Performance metrics included:
\begin{itemize}
\item	Mean Average Precision at IoU 0.5 (mAP@0.5)
\item	Mean Average Precision across IoU thresholds (mAP@0.5:0.95)
\item	Precision
\item	Recall
\item	F1-score
\end{itemize}

To ensure comparability, evaluation thresholds and confidence settings were kept identical across all models.

\begin{figure}[H]
\centering
\includegraphics[width=13.0 cm]{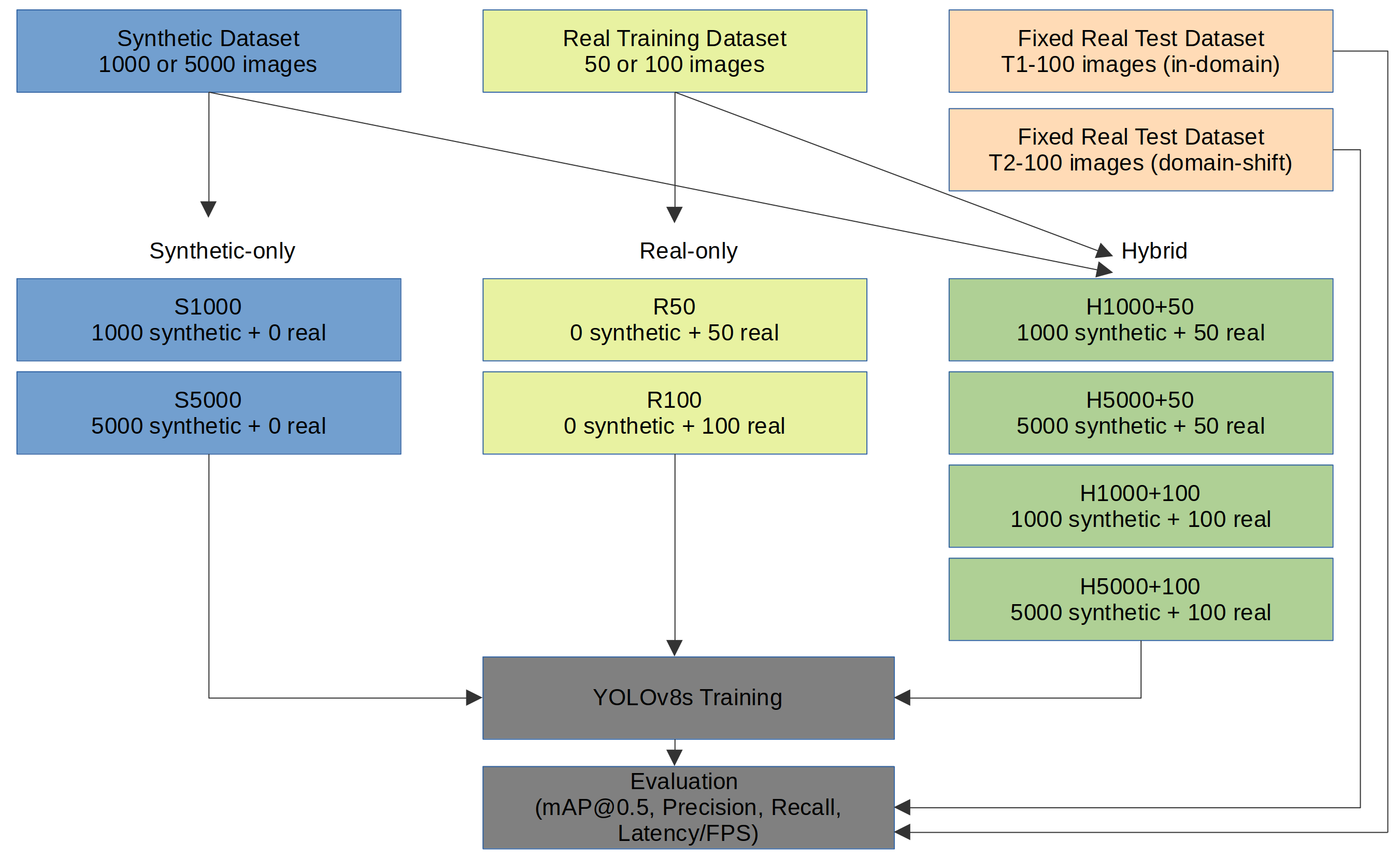}
\caption{Overview of the evaluated training regimes and common evaluation protocol. Synthetic-only, real-only, and hybrid models were trained 
using different combinations of synthetic and real training images, while all models were evaluated on the same fixed real-world test set.\label{fig3}}
\end{figure}

An overview of the training regimes and evaluation protocol is shown in Figure~\ref{fig3}

\unskip

\subsection{Deployment Preparation for Edge Inference}
\label{sec:edge_deployment}

After training, the best-performing models from each regime were exported to ONNX format and optimized using TensorRT for deployment on the 
Jetson Orin NX (16GB configuration). Optimization was performed using FP16 precision as the primary deployment mode.

For inference benchmarking:
\begin{itemize}
\item	Batch size was set to 1.
\item	Input resolution matched training resolution.
\item	End-to-end latency (including preprocessing, inference, and postprocessing) was measured.
\item	Power configuration and software stack versions were documented.
\end{itemize}

This unified pipeline ensured that both accuracy and deployment feasibility were evaluated under consistent and reproducible conditions.

This methodology enables a controlled quantitative analysis of how synthetic dataset size and hybrid training influence sim-to-real transfer 
performance and whether such improvements can be maintained under embedded edge deployment constraints.

\section{Results}

\subsection{Experimental Setup}

All models were evaluated on two fixed real-world test datasets, each consisting of 100 annotated images. 
The first dataset (T1-100) represents in-domain conditions, with plastic fruit and background settings consistent with the training data. 
The second dataset (T2-100) represents a domain shift scenario, containing real fruit and a different background not observed during training.
Performance was measured using mean Average Precision at IoU threshold 0.5 (mAP@0.5), precision, and recall. 

Training was performed using the Ultralytics framework with YOLO-based architectures.
For deployment experiments, trained models were exported to ONNX format and optimized using TensorRT. 
Inference benchmarking was conducted on a Jetson Orin NX 16GB embedded GPU platform.

\subsection{Detection Performance on Real Test Data}
Results of model performance on the in-domain test dataset (T1-100) are summarized in Table~\ref{tab1}. 
Models trained exclusively on real data achieve the highest performance, with mAP@0.5 exceeding 0.97. 
In contrast, models trained solely on synthetic data perform significantly worse, indicating a domain gap between synthetic and real observations. 

Hybrid training strategies, combining synthetic and real data, recover most of the performance, achieving results comparable to real-only 
training while requiring fewer real samples. 
Notably, increasing the amount of synthetic data alone does not lead to substantial improvements.

\begin{table}[H]
\centering
\caption{Detection performance on the real-world test dataset (T1-100), consisting of plastic fruit under the same background conditions as the training data.}
\label{tab1}
\begin{tabular}{lcccc}
\toprule
\textbf{Training Regime} & \textbf{mAP@0.5} & \textbf{mAP@0.5:0.95} & \textbf{Precision} & \textbf{Recall} \\
\midrule
Real (50 images)        & 0.978 & 0.875 & 0.968 & 0.959 \\
Real (100 images)       & 0.983 & 0.883 & 0.972 & 0.979 \\
Synthetic (1000)        & 0.626 & 0.493 & 0.487 & 0.713 \\
Synthetic (5000)        & 0.656 & 0.504 & 0.600 & 0.800 \\
Hybrid (1000 + 50)      & 0.974 & 0.844 & 0.949 & 0.965 \\
Hybrid (5000 + 50)      & 0.975 & 0.846 & 0.952 & 0.950 \\
Hybrid (1000 + 100)     & 0.983 & 0.877 & 0.961 & 0.982 \\
Hybrid (5000 + 100)     & 0.984 & 0.875 & 0.966 & 0.971 \\
\bottomrule
\end{tabular}
\end{table}

Results under domain shift conditions are presented in Table~\ref{tab2}. 
Compared to Table~\ref{tab1}, all models exhibit a noticeable drop in performance due to differences in object appearance and background. 
Models trained solely on synthetic data show limited generalization capability, while real-only models also degrade significantly.

Hybrid training improves robustness, with combinations of synthetic and real data achieving the best trade-off. 
However, increasing the amount of synthetic data alone does not consistently improve performance, highlighting the importance of real data for domain adaptation.

Interestingly, increasing the amount of real training data from 50 to 100 images does not improve performance under domain shift, suggesting potential overfitting to the training distribution.

\begin{table}[H]
\centering
\caption{Detection performance on the real-world test dataset (T2-100), consisting of real fruit and a different background not seen during training.}
\label{tab2}
\begin{tabular}{lcccc}
\toprule
\textbf{Training Regime} & \textbf{mAP@0.5} & \textbf{mAP@0.5:0.95} & \textbf{Precision} & \textbf{Recall} \\
\midrule
Real (50 images)        & 0.748 & 0.609 & 0.749 & 0.650 \\
Real (100 images)       & 0.703 & 0.562 & 0.780 & 0.610 \\
Synthetic (1000)        & 0.565 & 0.422 & 0.545 & 0.579 \\
Synthetic (5000)        & 0.510 & 0.359 & 0.667 & 0.530 \\
Hybrid (1000 + 50)      & 0.747 & 0.576 & 0.640 & 0.698 \\
Hybrid (5000 + 50)      & 0.737 & 0.569 & 0.794 & 0.679 \\
Hybrid (1000 + 100)     & 0.743 & 0.580 & 0.788 & 0.666 \\
Hybrid (5000 + 100)     & 0.680 & 0.527 & 0.799 & 0.612 \\
\bottomrule
\end{tabular}
\end{table}

Overall, the results indicate that synthetic data alone is not sufficient to fully replace real-world data in this application, 
but can substantially reduce the need for real annotations when used in a hybrid training strategy. 
Hybrid models achieve near real-data performance in in-domain scenarios and offer improved generalization under domain shift. 

Nevertheless, increasing the amount of synthetic data does not consistently lead to performance gains, highlighting limitations in capturing 
the variability of real-world appearances, particularly for natural objects such as fruit. 
This suggests that the quality and realism of synthetic data are more critical than its quantity.

\subsection{Quantitative and Qualitative Evaluation}

\begin{figure}[H]
\centering
\begin{subfigure}{0.48\textwidth}
\includegraphics[width=\linewidth]{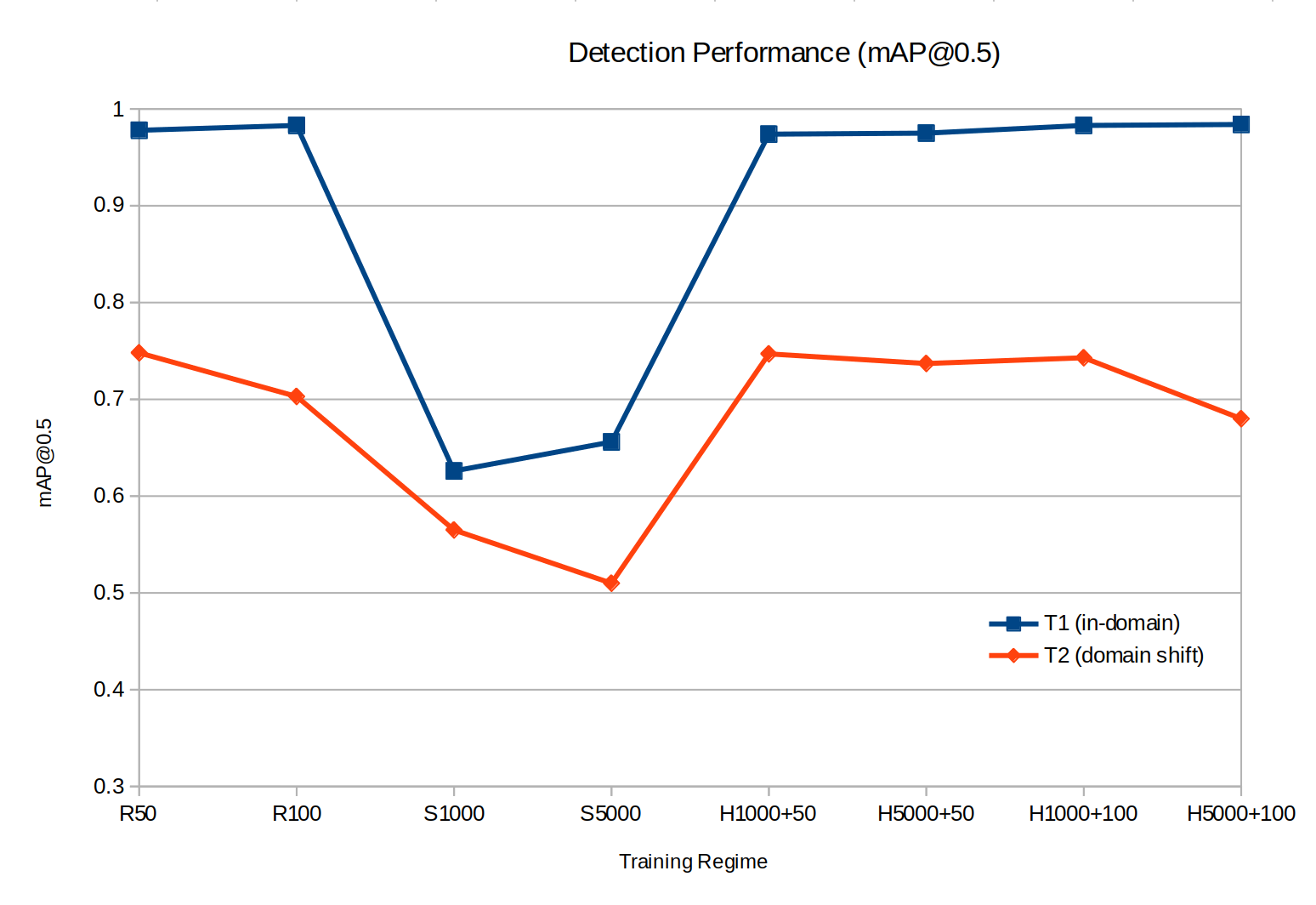}
\end{subfigure}
\begin{subfigure}{0.48\textwidth}
\includegraphics[width=\linewidth]{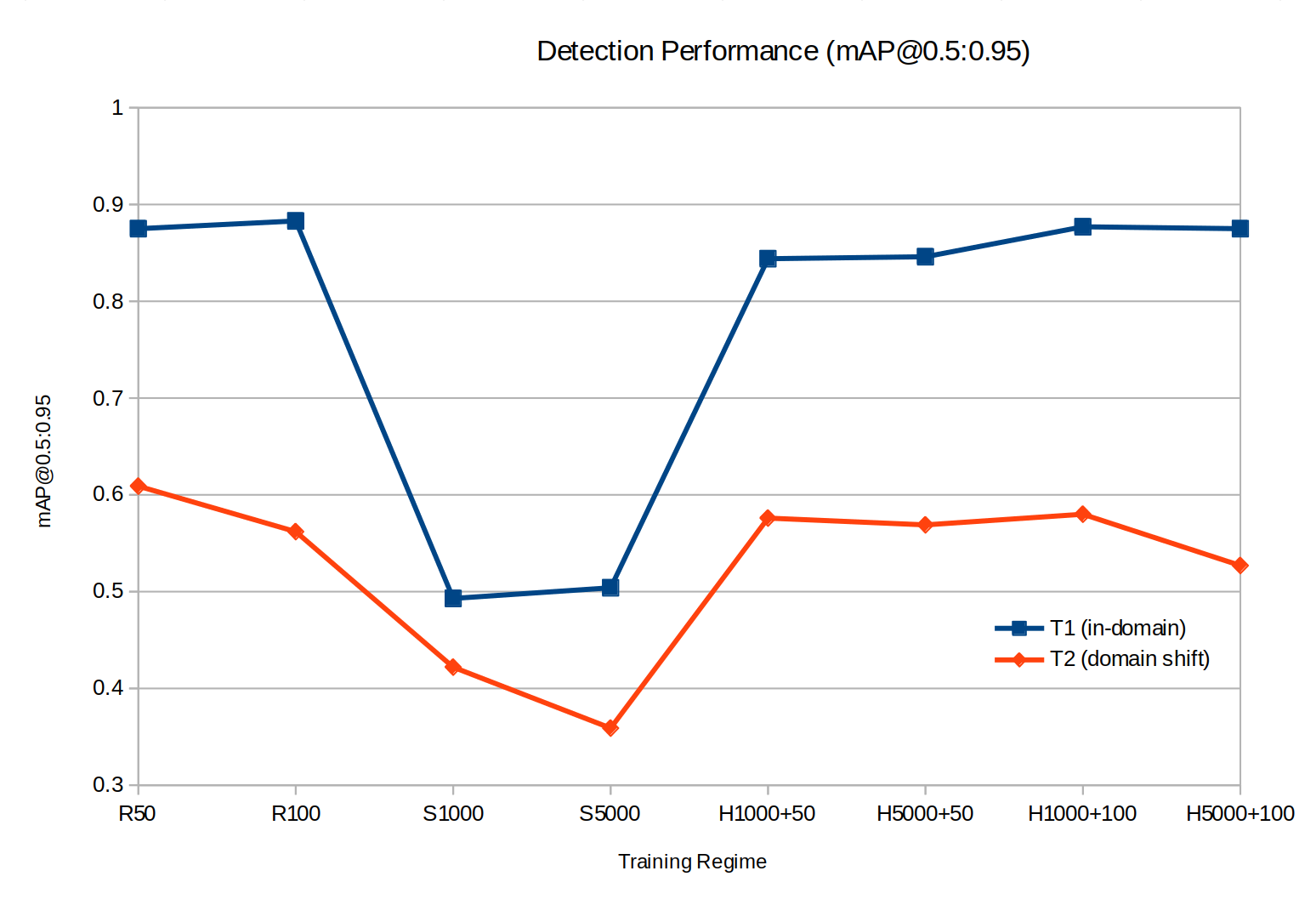}
\end{subfigure}
\caption{
(\textbf{a}) Detection performance (mAP@0.5) across different training regimes for in-domain (T1) and domain shift (T2) test datasets. 
The results show a significant performance drop under domain shift, while hybrid training improves robustness. 
(\textbf{b}) Detection performance (mAP@0.5:0.95) across different training regimes for in-domain (T1) and domain shift (T2) test datasets. 
The results confirm reduced performance under domain shift and highlight the benefit of hybrid training strategies.\label{fig4}}
\end{figure}

\begin{figure}[H]
\centering

\begin{subfigure}{0.48\textwidth}
\centering
\includegraphics[width=\linewidth]{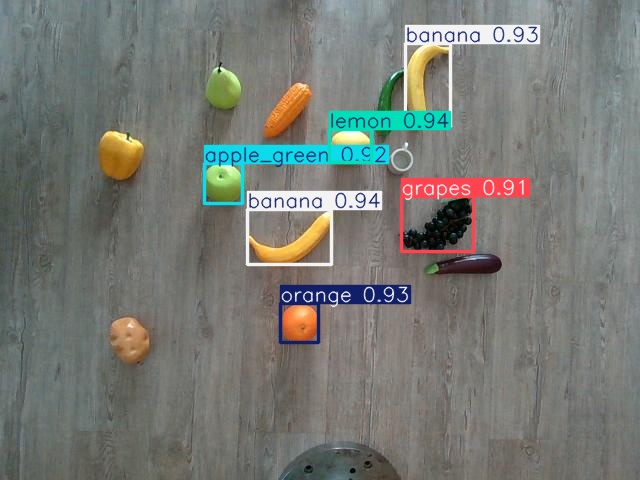}
\caption{}
\end{subfigure}
\hfill
\begin{subfigure}{0.48\textwidth}
\centering
\includegraphics[width=\linewidth]{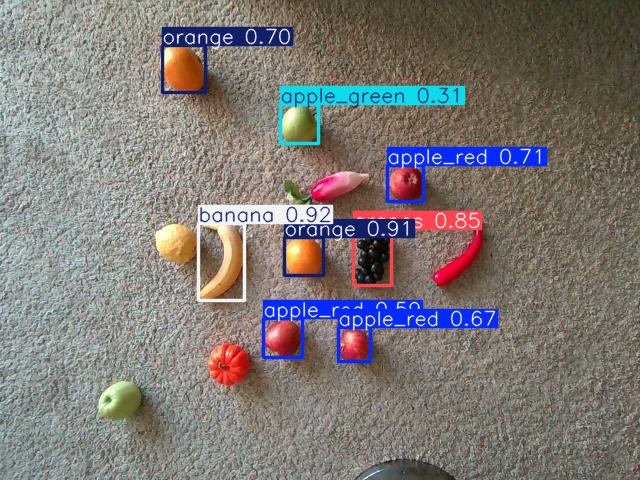}
\caption{}
\end{subfigure}

\vspace{0.3cm}

\begin{subfigure}{0.48\textwidth}
\centering
\includegraphics[width=\linewidth]{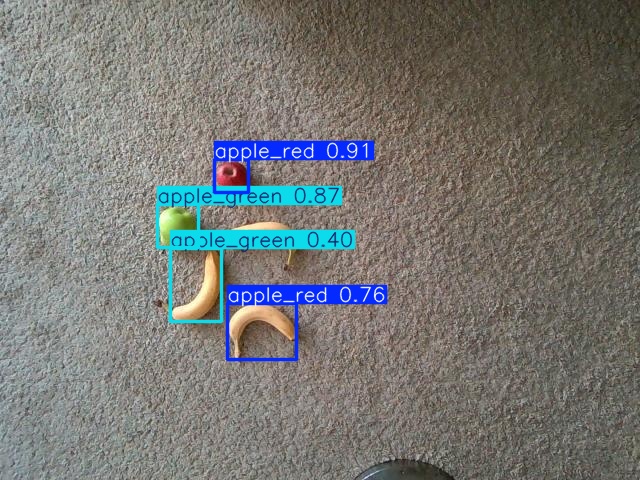}
\caption{}
\end{subfigure}
\hfill
\begin{subfigure}{0.48\textwidth}
\centering
\includegraphics[width=\linewidth]{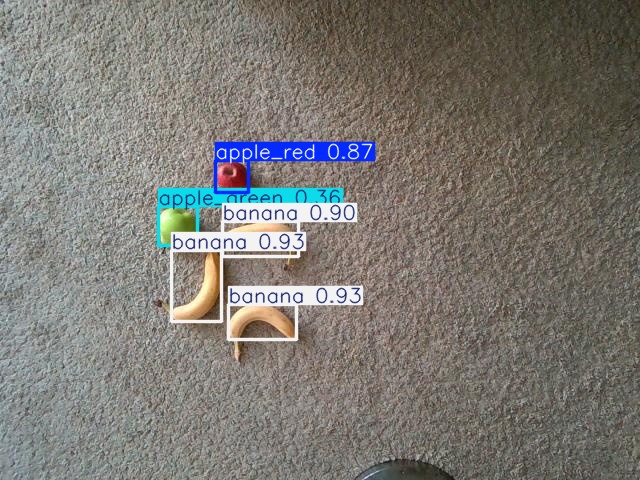}
\caption{}
\end{subfigure}

\caption{Qualitative detection results under different training regimes. 
(\textbf{a}) In-domain example using hybrid training (H1000+50) evaluated on T1-100. 
(\textbf{b}) Domain shift example using the same model evaluated on T2-100. 
(\textbf{c}) Failure case of a synthetic-only model (S1000) under domain shift. 
(\textbf{d}) Improved detection on the same scene using hybrid training (H1000+50).}
\label{fig5}

\end{figure}

To further analyze model behavior, both quantitative trends and qualitative detection results are presented. 
Figure~\ref{fig4} compares detection performance across real-only, synthetic-only, and hybrid training regimes under both in-domain 
and domain-shift test conditions, while Figure~\ref{fig5} provides representative 
detection examples under both in-domain and domain shift conditions.

Qualitative results in Figure~\ref{fig5} further highlight the differences between training regimes. 
Under in-domain conditions, all models achieve accurate detections with well-localized bounding boxes. 
However, under domain shift conditions, models trained solely on synthetic data exhibit missed detections and reduced localization accuracy. 

Hybrid models demonstrate more robust behavior, maintaining detection performance across varying object appearances and backgrounds.

\subsection{Comparison with YOLO26 Architecture}
This comparison is included as an exploratory extension and is limited to a single training regime and desktop evaluation.

To evaluate the impact of model architecture, selected experiments were repeated using the YOLO26s model. 
Evaluation was performed on the same test datasets under identical conditions, but only on a desktop GPU due to deployment constraints.

On the T2-100 dataset, YOLO26s trained on the R50 regime achieved mAP@0.5 of 0.804 and mAP@0.5:0.95 of 0.660, 
compared to 0.748 and 0.609 obtained with YOLOv8s. This corresponds to an improvement of approximately 7–8\% in detection performance (Table~\ref{tab3}).

\begin{table}[H]
\centering
\caption{Detection performance on the real-world test dataset (T2-100), consisting of real fruit and a different background not seen during training.}
\label{tab3}
\begin{tabular}{lccc}
\toprule
\textbf{Model} & \textbf{mAP@0.5} & \textbf{mAP@0.5:0.95} & \textbf{Precision / Recall} \\
\midrule
YOLOv8s (R50)  & 0.748 & 0.609 & 0.749 / 0.650 \\
YOLO26s (R50)  & \textbf{0.804} & \textbf{0.660} & 0.833 / 0.673 \\
\bottomrule
\end{tabular}
\end{table}

These results indicate that architectural improvements can partially mitigate domain shift effects. 
However, due to the lack of embedded deployment evaluation, no conclusions regarding embedded deployment performance can be drawn from this comparison.

\subsection{Embedded Inference Performance}

The trained YOLOv8s model was deployed on a Jetson Orin NX 16 GB using TensorRT FP16 optimization. The system achieved an average inference 
latency of 18.9 ms per image at a resolution of 832×832, corresponding to approximately 53 FPS for model inference only.

Including preprocessing and postprocessing, the end-to-end processing time was 35.4 ms per image (28 FPS).

Inference results are summarized in Table~\ref{tab4}.

\begin{table}[H]
\centering
\caption{Inference performance on Jetson Orin NX.}
\label{tab4}
\begin{tabular}{l C{2cm} C{2cm} C{2.8cm} C{2.5cm} C{3cm}}
\toprule
\textbf{Model} & \textbf{Input Size} & \textbf{Precision} & \textbf{Inference Latency (ms)} & \textbf{E2E Latency (ms)} & \textbf{FPS (Inference)/(E2E)} \\
\midrule
YOLOv8s & 832×832 & TensorRT FP16 & 18.9 & 35.4 & 53 / 28 \\
\bottomrule
\end{tabular}

\end{table}

These measurements follow the evaluation protocol described in \autoref{sec:edge_deployment}.

\section{Discussion}

This study investigated the transfer of synthetic data for sim-to-real effectiveness in object detection (fruit) under constrained real-data conditions. 
The results demonstrate that models, which were trained only on real data, achieve the best performance under in-domain conditions, while 
models trained purely on synthetic data perform significantly worse. This indicates a considerable domain gap between simulated and real environments.

However, combining synthetic and real data in hybrid training strategies consistently improves performance compared to synthetic-only models 
and approaches the performance of real-only training. This indicates that synthetic data can reduce the amount of required real-world annotations, 
although it cannot fully replace them.

Under the conditions of domain shift, all models exhibit a performance decrease, highlighting the sensitivity of learned representations to changes in 
object appearance and background. Hybrid models demonstrate improved robustness compared to synthetic-only approaches, but do not fully eliminate 
the domain gap.

From a deployment perspective, the optimized YOLOv8s model achieved real-time performance on the Jetson Orin NX, demonstrating that the proposed 
approach is feasible for embedded applications. The results indicate that it is possible to maintain competitive detection performance while 
meeting the latency constraints of edge devices.

It should be noted that larger model variants could potentially achieve higher detection accuracy. However, this study focuses on the YOLOv8s 
architecture as a representative lightweight model suitable for real-time inference on embedded hardware. This reflects a practical trade-off 
between accuracy and computational efficiency, which can be critical in robotics deployments and industrial applications.

It should also be noted that the difficulty of the detection task is influenced by the number and visual similarity of classes. A reduced class 
set would likely lead to improved performance due to decreased inter-class ambiguity and increased data per class. However, the chosen multi-class 
setup reflects a more realistic application scenario.

Overall, the results indicate that the synthetic data effectiveness is strongly dependent on data realism and its integration with real-world 
samples. While synthetic data provides a scalable source of labeled data, its benefits are limited for objects with complex textures and natural 
variability, such as fruit.

\section{Conclusion}
This work evaluated the role of synthetic data in sim-to-real object detection under constrained real-data conditions. 
The results indicate that while synthetic data alone does not achieve the same performance as real-world data, it can substantially 
reduce the need for manual annotation when used in hybrid training strategies. 
Real-time deployment on an embedded platform further demonstrates the practical feasibility of the approach, highlighting the importance 
of balancing data realism, model accuracy, and computational efficiency.
Unlike prior work, this study provides a controlled quantitative analysis under extremely limited real-data conditions, which are 
representative of practical industrial deployment scenarios.

\section{Further Work}

Further work can focus on an improvement of realism and variability of synthetic data, especially for objects with complex textures and non-uniform 
appearance, like fruit. Techniques such as advanced material modeling, improved lighting simulation, and domain randomization may help reduce the domain gap.

Additionally, extending the evaluation to structured industrial objects, such as manufactured parts with well-defined geometry and materials, 
could provide further insights into scenarios where synthetic data is more effective.

Another important direction is the evaluation of different model architectures and optimization strategies for edge deployment. In particular, 
benchmarking newer architectures under TensorRT optimization on embedded platforms would enable a more comprehensive analysis of accuracy–latency 
trade-offs.

Finally, integrating domain adaptation techniques and semi-supervised learning approaches may further improve generalization performance while 
reducing the reliance on large amounts of annotated real-world data.

\section*{Acknowledgments}
The author gratefully acknowledges YASKAWA Europe GmbH for providing access to the Autonomous Control Unit (ACU) of the MOTOMAN NEXT platform 
(including the Jetson Orin NX hardware), utilized in this work. 
The author also thanks YASKAWA Europe GmbH for supplying the plastic fruit used in the experimental dataset preparation.

The author further acknowledges the use of publicly available 3D models obtained from Sketchfab for synthetic dataset generation.

During the preparation of this manuscript, the author used ChatGPT (OpenAI) for language refinement and text editing. The author has reviewed and 
edited the output and takes full responsibility for the content of this publication.

\section*{Conflict of Interest}
The author is employed by YASKAWA Europe GmbH. The company provided hardware and materials used in this study but had no role in the design 
of the study; in the collection, analyses, or interpretation of data; in the writing of the manuscript; or in the decision to publish the results.



\section*{Abbreviations}
The following abbreviations are used in this manuscript:

\noindent 
\begin{tabular}{@{}ll}
ACU & Autonomous Control Unit \\
IoU & Intersection over Union \\
mAP@0.5 & Mean Average Precision at IoU = 0.5 \\
mAP@0.5:0.95 & Mean Average Precision averaged over IoU thresholds from 0.5 to 0.95 \\
FPS & Frames Per Second \\
FP16 & 16-bit Floating Point Precision \\
ONNX & Open Neural Network Exchange \\
\end{tabular}

\end{document}